\documentclass[10pt,twocolumn,letterpaper]{article}

\usepackage[pagenumbers]{cvpr} % To force page numbers, e.g. for an arXiv version
\usepackage{multirow}
\usepackage{amsmath}
\usepackage{colortbl}
\usepackage{float}

\definecolor{cvprblue}{rgb}{0.21,0.49,0.74}
\usepackage[pagebackref,breaklinks,colorlinks,allcolors=cvprblue]{hyperref}

\def\paperID{9309} % *** Enter the Paper ID here
\def\confName{CVPR}
\def\confYear{2026}

\title{Low-Level Dataset Distillation for Medical Image Enhancement}

\author{Fengzhi Xu$^1$, Ziyuan Yang$^2$, Mengyu Sun$^1$, Joey Tianyi Zhou$^3$, Yi Zhang$^1$\\
$^1$School of Cyber Science and Engineering, Sichuan University, China\\
$^2$The Chinese University of Hong Kong, Hong Kong SAR, China \\
$^3$Agency for Science, Technology and Research (A*STAR), Singapore\\
{\tt\small xufz@stu.scu.edu.cn, cziyuanyang@gmail.com, mysun2001999@163.com}\\ 
{\tt\small zhouty@a-star.edu.sg, yzhang@scu.edu.cn}
}

\begin{document}
\maketitle
\begin{abstract}
Medical image enhancement is clinically valuable, but existing methods require large-scale datasets to learn complex pixel-level mappings. However, the substantial training and storage costs associated with these datasets hinder their practical deployment.
While dataset distillation~(DD) can alleviate these burdens, existing methods mainly target high-level tasks, where multiple samples share the same label. This many-to-one mapping allows distilled data to capture shared semantics and achieve information compression. In contrast, low-level tasks involve a many-to-many mapping that requires pixel-level fidelity, making low-level DD an underdetermined problem, as a small distilled dataset cannot fully constrain the dense pixel-level mappings.
To address this, we propose the first low-level DD method for medical image enhancement.
We first leverage anatomical similarities across patients to construct the shared anatomical prior based on a representative patient, which serves as the initialization for the distilled data of different patients. This prior is then personalized for each patient using a Structure-Preserving Personalized Generation~(SPG) module, which integrates patient-specific anatomical information into the distilled dataset while preserving pixel-level fidelity. 
For different low-level tasks, the distilled data is used to construct task-specific high- and low-quality training pairs. Patient-specific knowledge is injected into the distilled data by aligning the gradients computed from networks trained on the distilled pairs with those from the corresponding patient’s raw data.
Notably, downstream users cannot access raw patient data. Instead, only a distilled dataset containing abstract training information is shared, which excludes patient-specific details and thus preserves privacy. Extensive experiments across diverse medical modalities and low-level tasks demonstrate the effectiveness of our method.
\end{abstract}
\section{Introduction}
\label{sec:intro}
Medical image enhancement tasks, such as super-resolution~\cite{li2021review, yang2023deep} and restoration~\cite{yang2025patient, kumar2025denoising}, 
aim to improve low-quality~(LQ) images into high-quality~(HQ) ones, thereby supporting accurate diagnosis and clinical decision-making. However, existing methods typically rely on large, high-resolution medical datasets to learn complex pixel-level mappings, and the associated training and storage costs hinder practical deployment~\cite{qiu2019efficient}.

\begin{figure}[t]
  \centering
  \includegraphics[width=\linewidth]{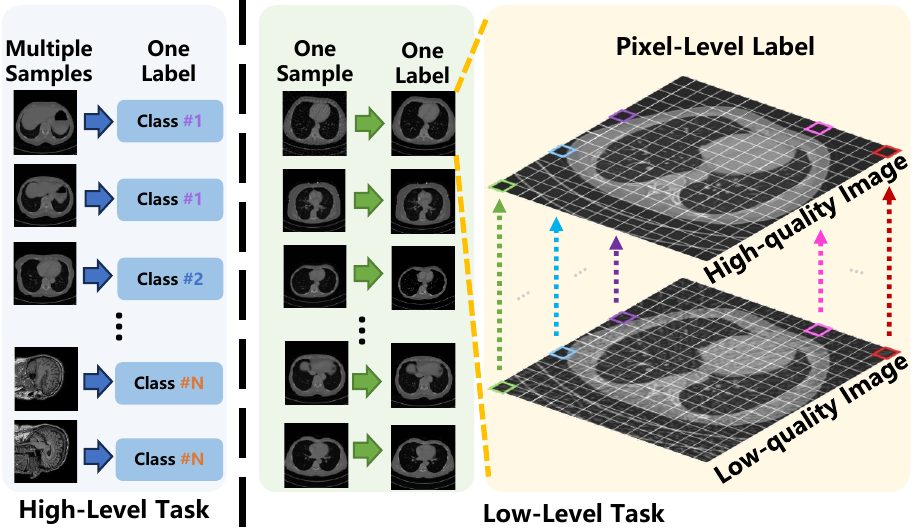}
   \caption{The comparison with the high-level and low-level tasks.}
   \label{fig:intro}
\end{figure}

To address these challenges, dataset distillation~(DD) has emerged as an efficient dataset compression paradigm that synthesizes a small distilled dataset while maintaining comparable training performance to the original large dataset~\cite{lei2023comprehensive, yu2023dataset}. 
While these methods improve data efficiency, they are primarily designed for high-level tasks such as classification, where multiple samples share the same label. This many-to-one mapping allows the distilled data to capture shared semantics among samples, making information compression feasible. However, low-level tasks involve a many-to-many mapping that requires pixel-level fidelity, which makes low-level DD underdetermined: the limited number of synthetic samples cannot adequately capture or constrain the dense pixel-level relationships between inputs and outputs. Fig.~\ref{fig:intro} clearly illustrates the difference between the two tasks.

To address this issue, we propose the first low-level DD method for medical image enhancement. Specifically, we first construct a shared anatomical prior from a representative patient, which serves as a common structural initialization for all patients. This design introduces global constraints that restrict the solution space and help mitigate the underdetermined nature of the problem.

Furthermore, considering the substantial anatomical variations among patients, we avoid enforcing cross-patient distillation. Instead, we focus on performing distillation individually for each patient to better capture patient-specific characteristics and preserve structural fidelity. This approach decomposes the highly underdetermined problem into a series of smaller and more manageable subproblems. 

Specifically, we design a Structure-Preserving Personalized Generation~(SPG) module to inject patient-specific anatomical and training information into the distilled data. Specifically, SPG modulates the shared anatomical prior with a learnable patient-specific code, enabling the patient-specific distilled data to adapt to individual anatomical structures. To preserve pixel-level fidelity, an image is randomly selected from the prior and fused with the patient-specific distilled data.
Given that different low-level tasks have distinct quality degradation models, the distilled data is mapped to task-specific low-quality representations for generating corresponding high- and low-quality data pairs.
Finally, patient-specific training knowledge is incorporated by aligning the gradients computed from the distilled data with those from the patient’s raw data, ensuring that the distilled dataset captures both anatomical structures and task-relevant information.

Notably, downstream users cannot access raw patient data. Instead, only a distilled dataset containing abstract training information is shared, which excludes patient-specific details and thus preserves privacy. Extensive experiments across different settings validate the effectiveness of our method on multiple medical modalities, including Computed Tomography (CT) and Magnetic Resonance Imaging~(MRI), as well as across various low-level tasks such as super-resolution and image restoration.
The main contributions of this paper are summarized as follows:
\begin{itemize}
    \item We propose a general DD method for low-level medical image enhancement, applicable across diverse modalities and tasks. To the best of our knowledge, this is the first attempt to explore DD in the low-level domain.
    \item We leverage anatomical similarities across patients to build a shared anatomical prior as initialization, while capturing patient-specific variations through patient-specific distillation.
    \item We introduce a patient-wise personalization module to embed patient- and task-specific knowledge into the distilled data by aligning gradients computed from both distilled and raw data.
\end{itemize}
\section{Related Work}
\label{sec:related}

\subsection{Medical Image Enhancement}
Medical image enhancement, including super-resolution~\cite{mahapatra2019image, lu2025information} for enhancing image details and restoration~\cite{yang2025patient, 11185185} for denoising, aims to recover high-quality images from degraded inputs while preserving critical details to support accurate diagnosis. With the rapid progress of deep learning in recent years, numerous breakthroughs have been achieved in medical image enhancement tasks across diverse medical imaging modalities~\cite{ahmad2022new, umirzakova2024medical}. However, these methods typically rely on large‐scale datasets, resulting in substantial storage and computational overhead, which limits the development and application of such approaches.

\begin{figure*}[!t]
\centering
\includegraphics[width=\textwidth]{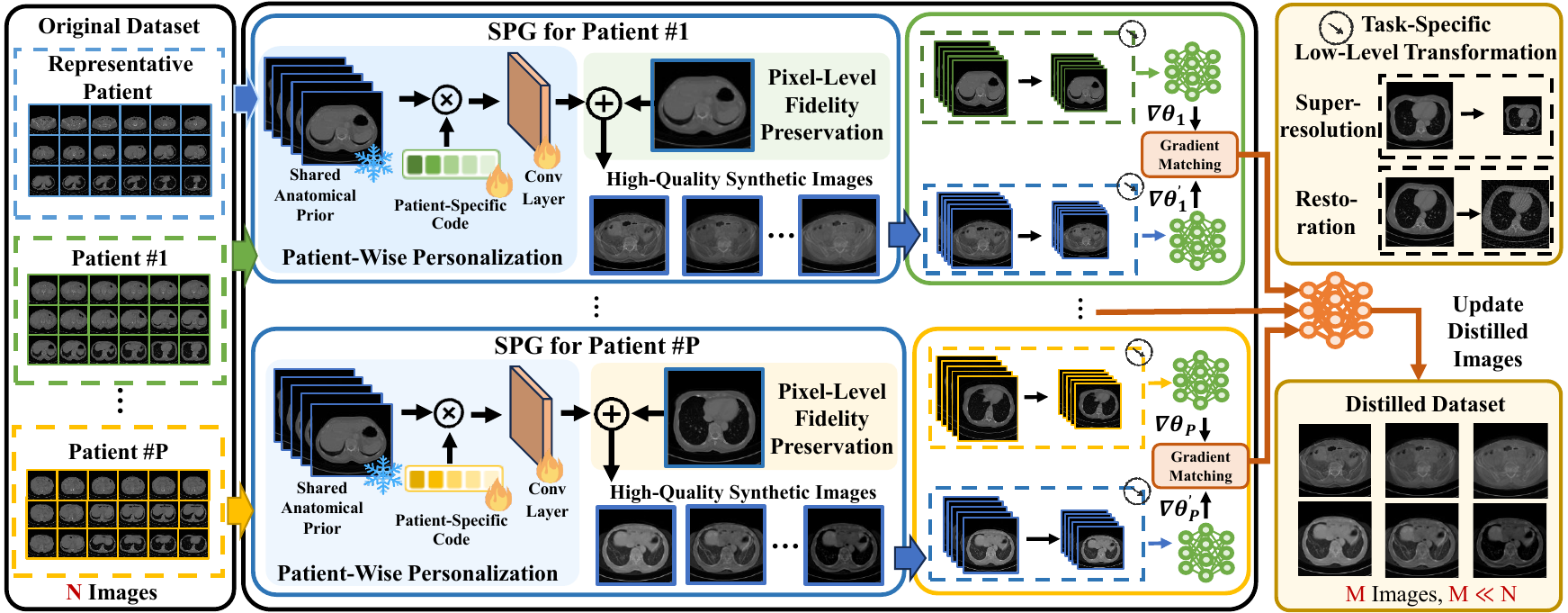} 
\caption{The overview of our proposed method.}
\label{fig:framework}
\end{figure*}

\subsection{Dataset Distillation}
DD has emerged as an efficient data compression paradigm that synthesizes a small distilled dataset while maintaining comparable training performance to the original large dataset~\cite{lei2023comprehensive, yu2023dataset}. 
DD methods can be classified according to different matching strategies as follows: 1) Performance matching, which formulates distillation as a bi-level optimization problem~\cite{wang2018dataset}, with improvements such as momentum-enhanced inner-loop updates~\cite{deng2022remember} and kernel-based approximations~\cite{nguyen2020dataset, nguyen2021dataset, loo2022efficient}. 2) Parameter matching, which aligns model optimization trajectories via gradient matching~\cite{zhao2021dataset} or multi-step updates~\cite{cazenavette2022dataset}, often integrated with strategies like symmetric augmentation~\cite{zhao2021datasetdsa} or learnable soft labels~\cite{cui2023scaling}. 3) Distribution matching, which aligns feature distributions between synthetic and real data, such as through class-wise~\cite{zhao2023datasetdm} or layer-wise statistical~\cite{wang2022cafe} alignment. However, these methods are primarily tailored for high-level tasks, making them ineffective for low-level tasks.
\section{Problem Statement}
\label{sec:prob_state}
Current DD methods mainly focus on high-level tasks, such as classification, while existing techniques struggle to generalize to low-level tasks. To clearly articulate the underlying challenges, this section formally defines DD for both high- and low-level tasks, highlights their key distinctions, and reveals the inherent difficulty of low-level DD.

Given a large high-level classification dataset $\mathcal{T}^h = \{\mathbf{x}_n, y_n\}_{n=1}^N (y_n\in \{1, 2, ..., C\})$ with $C$ class, DD seeks to extract the knowledge of $\mathcal{T}^h$ into a compact synthetic dataset $\mathcal{S}^h=\{\mathbf{\tilde{x}}_m, \tilde{y}_m\}_{m=1}^M(\tilde{y}_m\in \{1, 2, ..., C\})$, and $M\ll N$. DD seeks to train a model parameterized by $\theta_{\mathcal{S}^h}$ on $\mathcal{S}^h$ such that it achieves performance comparable to that of a model with parameters $\theta_{\mathcal{T}^h}$ trained on $\mathcal{T}^h$. This goal can be formulated as:
\begin{equation}
    \mathbb{E}_{(\mathbf{x},y)\sim \mathcal{P}_{RDD}}|l^h(\psi(\mathbf{x}; \theta_{\mathcal{T}^h}), y) - l^h(\psi(\mathbf{x}; \theta_{\mathcal{S}^h}), y)| \leq \epsilon,
\end{equation}
where $\mathcal{P}_{RDD}$ represents the real data distribution, $\epsilon$ is a small tolerance value, and $l^h$ and $\psi$ represent the loss function and the network for high-level tasks, respectively.

In high-level tasks, multiple samples often share the same label, forming a many-to-one mapping. This mapping allows synthetic data to focus on capturing the common high-level semantics within each class, which makes high-level DD feasible. In contrast, low-level tasks involve a many-to-many mapping between samples and labels and demand pixel-level fidelity, which substantially increases the difficulty of distillation.

For low-level tasks, we define the real dataset $\mathcal{T}^l=\{\mathbf{x}_n, \mathbf{y}_n\}_{n=1}^N (\mathbf{y}_n\in \mathbb{R}^{h\times w})$ and the synthetic dataset $\mathcal{S}^l=\{\mathbf{\tilde{x}}_m, \mathbf{\tilde{y}}_m\}_{m=1}^M (\mathbf{\tilde{y}}_m\in \mathbb{R}^{h\times w})$.
Based on these definitions, the objective of low-level DD is formulated as follows:
\begin{equation}
\label{eq:forde}
    \mathbb{E}_{(\mathbf{x},\mathbf{y})\sim \mathcal{P}_{RDD}}|l^l(\phi(\mathbf{x}; \theta_{\mathcal{T}^l}), \mathbf{y}) - l^l(\phi(\mathbf{x}; \theta_{\mathcal{S}^l}), \mathbf{y})| \leq \epsilon,
\end{equation}
where $\phi(\mathbf{x}; \theta_{\mathcal{T}^l})\in \mathbb{R}^{h\times w}$ represents a dense mapping output, such as a restored image, rather than a single class label as in high-level tasks. $l^l$ and $\phi$ denote the loss function and the network for low-level tasks, respectively.

From the above formulation, a network trained on the synthetic dataset must capture the pixel-level correspondence in the real data. However, under low-level tasks, the dense mapping between samples and labels makes it challenging for a small synthetic dataset to fully capture these relationships.
\section{Method}
\label{sec:met}
\subsection{Overview}
\label{sec:overview}
This paper proposes a low-level DD method for medical image enhancement, which aims to synthesize a compact dataset that maintains training performance comparable to the raw dataset. The overall framework is illustrated in Fig.~\ref{fig:framework}.
Considering the anatomical similarity among patients, we construct a shared anatomical prior from a representative patient to serve as the initialization. Then, we propose an SPG module to inject patient-specific anatomical and training information into the prior. Since image degradation models differ across low-level tasks, we design task-specific low-level transformations to simulate low- and high-quality image pairs based on the distilled data. Finally, the distilled and real pairs are used to train the image enhancement network, and their gradients are matched to update the previous module. 

\subsection{Structure-Preserving Personalized Generation}
\label{sec:ppm}
To generate structurally faithful distilled data, we propose a patient-specific generation pipeline. Our proposed SPG pipeline consists of three steps: initialization, patient-wise personalization, and low- and high-quality image pair generation. The details of each stage are elaborated in the following sections.

\noindent \textbf{\textit{Initialization.}} As introduced earlier, high-level tasks aim to construct a many-to-one mapping, where different images correspond to a shared label. In contrast, low-level tasks involve a different mapping, as each image corresponds to a unique ground truth, and no multiple inputs share the same target. Consequently, the distilled data is expected to capture many-to-many mapping knowledge within a limited dataset, making low-level DD an underdetermined problem.

To alleviate this issue, we leverage the fact that different patients share similar anatomical structures and introduce a shared anatomical prior as initialization. This prior provides global constraints that restrict the solution space and help mitigate the underdetermined nature of the problem.
Specifically, the shared anatomical prior is constructed by randomly selecting $n$ slices from the 3D volume (CT or MRI) of a single representative patient in the raw dataset, denoted as $U\in \mathbb{R}^{v\times h\times w}$, where $v$ represents the number of randomly selected images~(NRI). These selected slices serve as the shared initialization for subsequent distillation between different patients.

\noindent \textbf{\textit{Patient-Wise Personalization.}} After initialization, our goal is to inject information from different patients into the distilled dataset. However, patients differ in their anatomical composition, so forcing a single distilled sample to represent multiple patients is infeasible and would degrade the quality of the distilled dataset. Therefore, our method avoids across-patient distillation, ensuring that the distilled data preserves patient-specific anatomical and training information.

To achieve this, we propose a patient-wise personalization module, which includes two components: a learnable patient-specific code $d_p\in \mathbb{R}^q$ for the $p$-th patient and a patient-agnostic convolutional layer paramized by $\theta^c$.

\begin{figure}[!t]
  \centering
  \includegraphics[width=0.75\linewidth]{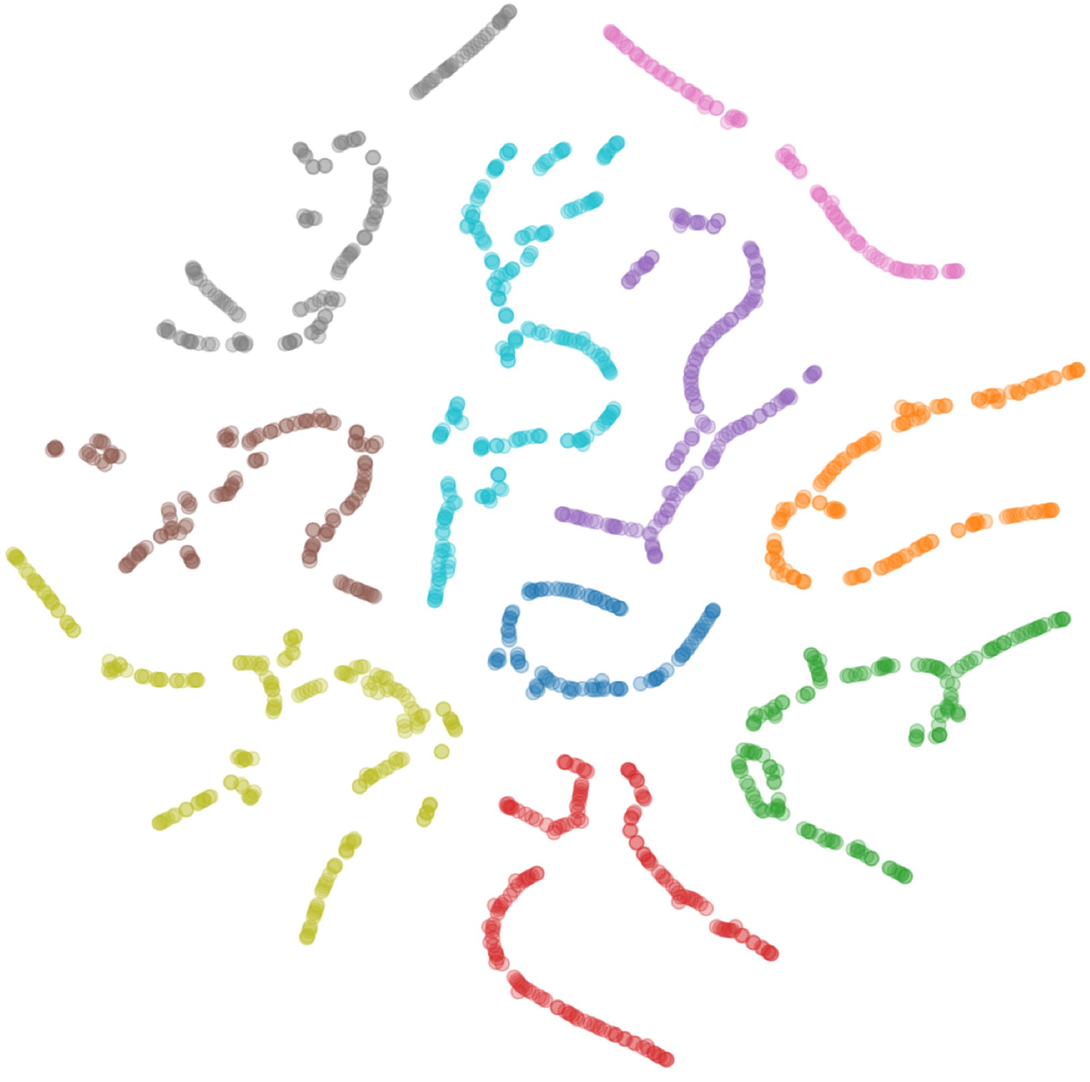}
   \caption{t-SNE visualization of gradient data from different patients. Different colors denote different patients.}
   \label{fig:tsne_g}
\end{figure}

Specifically, we adjust the shared anatomical prior $U$ using learnable patient-specific code $d_p$, which can be formulated as follows:
\begin{equation}
    f(U, d_p) = \text{Concat}(d_p^1 \cdot U,...,d_p^q \cdot U),
\end{equation}
where $f(\cdot)$ denotes the adjustment operation, and its output lies in $\mathbb{R}^{v\times q \times h \times w}$. $d_p^q$ represents the $q$-th element of $d_p$.

The adjusted output is then fed into a patient-agnostic convolutional layer to match its dimension to the number of images per patient~(IPP), denoted as $i$. Finally, the personalized data has a dimension of $\mathbb{R}^{i \times h \times w}$.

\noindent \textbf{\textit{Pixel-Level Fidelity Prservation.}} Since pixel-level fidelity is important in low-level tasks, we introduce a pixel-level fidelity preservation step to maintain this fidelity. Specifically, for the $p$-th patient, a slice $\mathbf{u}_p\in \mathbb{R}^{h\times w}$ is randomly selected from $U$ and duplicated $i$ times to match the convolutional output dimension. The duplicated slice is then added to the personalized data. Then, the overall process of SPG can be formulated as follows:
\begin{equation}
    \mathbf{\tilde{Y}_p} = \text{Conv}(f(U, d_p); \theta^c)+\mathbf{u}_p
\end{equation}
where $\mathbf{\tilde{Y}_p}\in \mathbb{R}^{i\times h\times w}$ denotes the distilled data for the $p$-th patient.

\noindent \textbf{\textit{Low- and High-Quality Image Pair Generation.}} Different low-level tasks follow distinct degradation models, so we first construct task-specific pairs of low- and high-quality images for subsequent training.
Specifically, the personalized data is treated as the high-quality data, and the pair construction process based on the personalized data and the real high-quality data can be formulated as:
\begin{equation}
    \mathcal{S}_p^l = \{D(\mathbf{\tilde{y}}_m), \mathbf{\tilde{y}}_m\}_{m=1}^{i}, \mathcal{T}_p^l = \{D(\mathbf{y}_n^p), \mathbf{y}_n^p\}_{n=1}^{N_p},
\end{equation}
where $\mathbf{\tilde{y}}_m \in \mathbb{R}^{h\times w}$ denotes the personalized data sample from $\mathbf{\tilde{Y}_p}$.
$y_n^p \in \mathbb{R}^{h \times w}$ is the real high-quality image from the $p$-th patient, and $N_p$ represents the number of samples. $D(\cdot)$ denotes a degradation operation that simulates low-quality images, such as resizing for super-resolution or adding noise in the measurement data for medical image restoration.

\subsection{Patient-Awareness Gradient Matching}
\label{sec:gm}
Due to patients differ in their anatomical composition, cross-patient distillation may damage individual structural characteristics and prevent the distilled data from converging effectively. Therefore, we split the dense mapping into multiple sub-mappings and perform distillation independently for each patient, which prevents training collapse and better preserves patient-specific anatomical information.

Specifically, our goal is to achieve patient-aware dataset distillation, which compresses each patient’s large dataset $\mathcal{T}_p^l$ into a smaller dataset $\mathcal{S}_p^l$, with the collection of all $\mathcal{S}_p^l$ forming the overall distilled dataset $\mathcal{S}^l$. We define a raw dataset $\mathcal{T}^l$ containing multiple medical imaging data of $P$ patients. 
Our optimization objective is to ensure that the parameters learned from the distilled dataset are close to those learned from the raw dataset. Accordingly, the objective mentioned in Eq.~\eqref{eq:forde} can be reformulated as follows:
\begin{equation}
     \min\limits_{\mathcal{S}^l}\sum\limits_{p=1}^{P} \mathcal{L}_{PM}(\theta_{\mathcal{S}_p^l},\theta_{\mathcal{T}_p^l}),
\end{equation}
where $\mathcal{L}_{PM}$ denotes the parameter matching function, and $\theta_{\mathcal{S}_p^l}$ and $\theta_{\mathcal{T}_p^l}$ represent the network parameters learned from the synthetic dataset and the real dataset of the $p$-th patient.

We aim to match the gradients computed from both the raw and distilled data. However, injecting the training information from multiple patients into a single distilled data is challenging due to substantial anatomical differences across patients. To stand this, we visualize the gradient data from different patients using t-SNE~\cite{maaten2008visualizing}, and the results are illustrated in Fig.~\ref{fig:tsne_g}. It can be observed that gradients trained on different patients are discriminative, while those trained on the same patient are more similar.

Therefore, in this paper, we propose a patient-specific distillation process to prevent the cross-patient information injection, which could lead to training collapse.
Specifically, the patient-awareness gradient matching process can be formulated as follows:
\begin{equation}
\mathcal{L}_{PGM}(\nabla \theta_p, \nabla \theta_p') = 1- \frac{\nabla \theta_p'  \cdot \nabla \theta_p}{ ||\nabla \theta_p'||||\nabla \theta_p||},
\end{equation}
where $\nabla \theta_p$ and $\nabla \theta_p'$ denote the gradients computed on the real data $\mathcal{T}^l_p$ and the synthetic data $\mathcal{S}^l_p$ for the $p$-th patient, respectively.

For computing the gradients used in the distillation, we employ the commonly used mean squared error~(MSE) as the task loss to measure the pixel-level fidelity between the output of the network and the corresponding high-quality image. The loss function can be formulated as follows:
\begin{equation}
  l^{l}_{MSE}(\mathbf{x}, \mathbf{y}) = ||\phi(D(\mathbf{y});\theta)-\mathbf{y}||^2,
  \label{eq:llloss}
\end{equation}
where $\phi_{\theta}(D(\mathbf{y}))$ is the enhanced image and $\mathbf{y}$ is the ground-truth from the real dataset. For the synthetic dataset, the loss is formulated analogously, with the paired data replaced by synthetic data $\mathcal{S}_p^l$.

In this way, we obtain a distilled low-level dataset that is significantly smaller than the raw dataset, while still preserving essential information.

\subsection{Downstream Task Usage}
\label{downstream}
Medical data are typically large and sensitive, and cannot be directly shared due to privacy regulations and ethical concerns.
To enable efficient training while preserving privacy, we share the distilled dataset $\mathcal{S}^l=\{\mathbf{\tilde{x}}_m, \mathbf{\tilde{y}}_m\}_{m=1}^M$ with downstream users.

Based on this synthetic dataset, downstream users utilize paired high- and low-quality data to efficiently optimize image enhancement networks using the following formulation:
\begin{equation}
  \phi_{\theta_{\mathcal{S}^l}} = \arg\min\limits_{\theta} ||\phi(\mathbf{\tilde{x}}; \theta)-\mathbf{\mathbf{\tilde{y}}}||^2,
  \label{eq:llloss}
\end{equation}
where $\mathbf{\tilde{x}}$ and $\mathbf{\tilde{y}}$ denote the paired synthetic data from the synthetic dataset $\mathcal{S}^l$.

In our pipeline, patients' raw data are never shared with users. Instead, only the distilled dataset is accessed, which encapsulates each patient’s informative knowledge. This enables the distilled dataset to retain valuable patient-specific information while preventing exposure of identifiable data, thereby ensuring privacy throughout the distillation process. The corresponding visualization is provided in Sec.~\ref{exp:ablation}. Moreover, since the distilled dataset is much smaller than the raw dataset, training efficiency is greatly improved. Notably, our method is task-agnostic. By modifying the degradation model $D(\cdot)$, it can be readily extended to different low-level medical image enhancement tasks.
\section{Experiment}
\label{sec:exp}

\begin{table*}[t!]
  \centering
  \small
  \caption{Quantitative super-resolution results of different methods trained on SRCNN with different modalities.}
  \begin{tabular}{llllll|llll}
    \toprule
    & & \multicolumn{4}{c}{CT (x4)} & \multicolumn{4}{c}{MRI (x2)}\\
    & & \multicolumn{2}{c}{NRI=5} & \multicolumn{2}{c}{NRI=10} & \multicolumn{2}{c}{NRI=5} & \multicolumn{2}{c}{NRI=10} \\
    \cmidrule(r){3-6} \cmidrule(r){7-10}
    \multicolumn{2}{c}{Methods} & PSNR~(dB)$\uparrow$ & SSIM$\uparrow$ & PSNR~(dB)$\uparrow$ & SSIM$\uparrow$ & PSNR~(dB)$\uparrow$ & SSIM$\uparrow$ & PSNR~(dB)$\uparrow$ & SSIM$\uparrow$\\
    \midrule
    \multicolumn{2}{c}{Full Data} & \multicolumn{2}{c}{PSNR=36.10±0.02} & \multicolumn{2}{c}{SSIM=94.86±0.03} & \multicolumn{2}{c}{PSNR=29.07±0.00} & \multicolumn{2}{c}{SSIM=90.46±0.02}\\
    \midrule
    \multirow{5}{*}{Base} & Random & 31.05±0.35 & 87.38±1.32 & 32.66±0.41 & 91.06±0.51 & 25.54±0.08 & 79.27±0.42 & 26.80±0.11  & 83.63±0.55\\
     & Random* & 31.37±0.36 & 87.98±0.67 & 32.76±0.19 & 90.78±0.69 & 25.70±0.13 & 79.87±0.57 & 26.74±0.13 & 83.54±0.44\\
     & Uniform & 31.39±0.24 & 87.84±0.98 & 32.46±0.28 & 90.67±0.26 & 25.94±0.06 & 80.80±0.25 & 26.77±0.06 & 83.79±0.23\\
     & Herding & 31.41±0.23 & 88.80±0.40 & 32.86±0.26 & 91.18±0.48 & 25.63±0.13 & 79.57±0.58 & 26.72±0.09 & 83.62±0.36\\
     & K-Center & 31.59±0.24 & 89.22±0.40 & 32.92±0.23 & 91.34±0.28 & 25.70±0.09 & 79.97±0.50 & 26.76±0.10 & 83.70±0.36\\
    \midrule
    \rowcolor{gray!12}
    & IPP=1 & 32.73±0.22 & 90.50±0.78 & 32.93±0.28 & 90.15±1.90 & 26.95±0.21 & 71.78±15.14 & 27.02±0.28 & 77.04±8.22\\
    \rowcolor{gray!12}\multirow{-2}{*}{Ours} & IPP=5 & \textbf{34.45±0.20} & \textbf{92.07±0.93} & \textbf{34.57±0.14} & \textbf{92.83±0.28} & \textbf{28.16±0.18} & \textbf{87.73±0.44} & \textbf{28.35±0.10} & \textbf{88.27±0.33}\\
    \bottomrule
  \end{tabular}
  \label{tab:compar_results}
\end{table*}

\label{exp:comparsion}
\begin{figure*}[h]
  \centering
  \includegraphics[width=\linewidth]{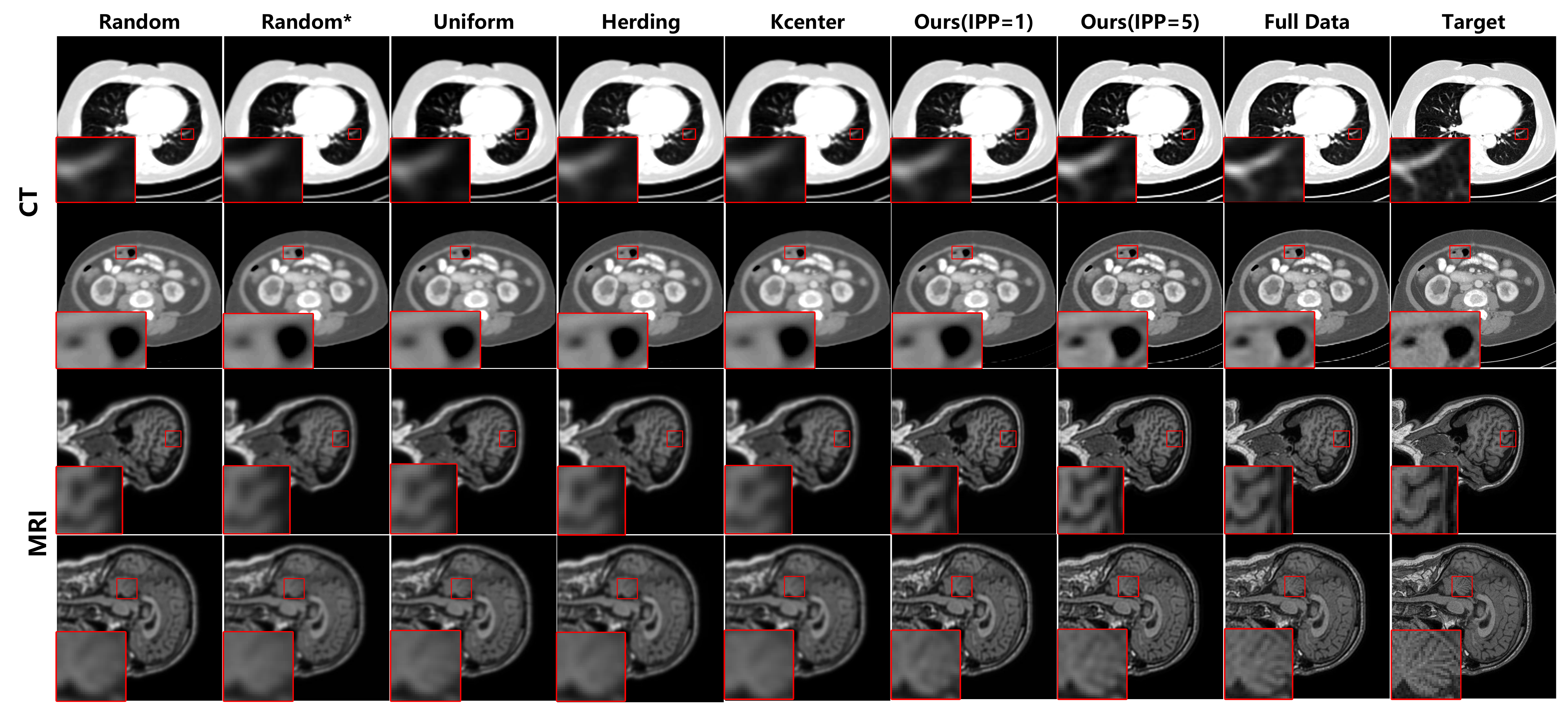}
   \caption{Qualitative super-resolution results of different algorithms with CT and MRI modality. The display window for the first row is [-950, 50] HU, while for the second row is [-290, 310] HU.}
   \label{fig:par}
\end{figure*}

\subsection{Experimental Setup}
\label{exp:setup}

\noindent \textbf{Implementation Details.} All experiments are conducted using PyTorch on an NVIDIA RTX 3090 GPU. We perform iterative distillation for 2,000 optimization steps.
SRCNN~\cite{dong2015image} and REDCNN~\cite{chen2017low}, two widely used baseline methods, are employed for the super-resolution and restoration tasks, respectively.
For our method, three independent distilled datasets are generated in each experiment, and five training–testing runs are conducted per dataset, resulting in 15 test scores whose mean and variance are reported. For CT super-resolution and CT restoration tasks, all networks are trained for 300 epochs, while MRI super-resolution networks are trained for 600 epochs.
For coreset-based baselines, after selecting the subset, we perform multiple independent training–testing runs on the selected data and report the mean and variance of the resulting test scores.\footnote{The code will be made publicly available for reproducibility.}

\noindent \textbf{Datasets.} We evaluate the proposed method on multiple imaging modalities, including CT and MRI.
For CT, we employ the public “NIH-AAPM-Mayo Clinic Low-Dose CT Grand Challenge”\footnote{The Mayo dataset link is https://www.aapm.org/grandchallenge/lowdosect/} dataset~\cite{mccollough2016tu}. In the super-resolution task, 50 images per patient from 10 patients are used for training, while in the restoration task, all images from 10 patients are used for training. For both tasks, data from two additional patients are used for testing.

For MRI, Calgary-Campinas-359 dataset\footnote{The CC-359 dataset link is https://www.ccdataset.com/}~\cite{souza2018open} is employed in our experiments. In the super-resolution task, 100 high-resolution MRI images per patient from 10 patients are used for training, and data from two additional patients are used for testing.

\noindent \textbf{Data Simulation.} We validate our method on two common medical image enhancement tasks—super-resolution and image restoration.
For the super-resolution task, the degradation operator reduces the 512$\times$512 CT images by a factor of $4$, while MRI images are first resized to 256$\times$256 and then downsampled by a factor of $2$ to generate low-quality inputs. 
For the low-dose CT~(LDCT) restoration task, the degradation is performed by adding noise to the projections followed by back-projection, with the photon count set to $10^4$.

\noindent \textbf{Comparison Methods.} We compare our method to five baseline methods: Random, Random*, Uniform, Herding, and K-Center. In the Random method, a subset of images is randomly selected from the entire dataset for training, while Random* samples images from a single patient. The Uniform method selects samples at equal intervals across all patients. The Herding method~\cite{herding} minimizes the distance between the coreset and dataset centers in feature space, and K-Center Greedy~(K-Center)~\cite{farahani2009facility, sener2018active} solves a minimax facility location problem.

\noindent \textbf{Evaluation Metrics.} We adopt Peak Signal-to-Noise Ratio~(PSNR) and Structural Similarity Index~(SSIM) as evaluation metrics. For both measures, higher scores reflect superior perceptual fidelity and structural integrity of the reconstructed images. 
Since low-level tasks lack explicit class concepts, we use the number of randomly selected real images (NRI) as a measure of dataset size.

\subsection{Comparison with Other Methods}
\noindent \textbf{Super-Resolution in the CT Modality.} For the CT modality, quantitative results are shown in Tab.~\ref{tab:compar_results}. Overall, coreset selection methods achieve similar performance at the same NRI, with Herding and K-Center demonstrating slightly better effectiveness and stability. 
Compared to these methods, our proposed method consistently outperforms them across all settings, clearly validating its effectiveness, and achieves the best performance when IPP=5. In our method, increasing IPP leads to noticeable improvements in performance, while also yielding more stable results.
As NRI increases, incorporating more information, all methods exhibit performance improvements. Ultimately, our proposed method achieves the best super-resolution results, with a PSNR of 34.57 dB and an SSIM of 92.83, demonstrating our effectiveness.
The qualitative results are shown in Fig~\ref{fig:par}. It can be noticed that our method significantly outperforms existing approaches on the CT super-resolution task, recovering anatomical details that more closely resemble the targets.

\noindent \textbf{Super-Resolution in the MRI Modality.} For the MRI modality, quantitative results, asillustrated in Tab.~\ref{tab:compar_results}, show a trend similar to that observed for CT. The performance of coreset selection methods is comparable, while our method consistently outperforms them and achieves the best results. As NRI increases, all methods improve. Remarkably, our approach reaches up to 97\% of the PSNR and SSIM performance obtained with the full dataset. The reason lies in that our method effectively injects patient-specific anatomical and training information into the distilled dataset, as described in the previous section.
In terms of qualitative results, utilizing the synthetic samples distilled by our approach leads to images with improved visual fidelity and anatomical detail, showing a closer resemblance to the target images, as shown in Fig.~\ref{fig:par}.

\subsection{Generalization Evaluation}
\label{exp:generalization}
\begin{table}[!t]
  \centering
  \small
  \caption{Quantitative super-resolution results using SRCNN distillation across different network architectures for the CT modality.}
  \resizebox{\columnwidth}{!}{
  \begin{tabular}{llllllll}
    \toprule
     & & \multicolumn{2}{c}{SRCNN} & \multicolumn{2}{c}{EDSR} & \multicolumn{2}{c}{SCTSRN}\\
    \cmidrule(r){3-4} \cmidrule(r){5-6} \cmidrule(r){7-8}
    \multicolumn{2}{c}{Algorithm} & PSNR~(dB)$\uparrow$ & SSIM$\uparrow$ & PSNR~(dB)$\uparrow$ & SSIM$\uparrow$ & PSNR~(dB)$\uparrow$ & SSIM$\uparrow$\\
    \midrule
    \multicolumn{2}{c}{Full data}& 36.10±0.02 & 94.86±0.03 & 38.29±0.03 & 96.19±0.01 & 35.01±0.09 & 94.68±0.07\\
    \midrule
    \multirow{4}{*}{Base} & Random & 32.66±0.41 & 91.06±0.51 & 34.14±0.04 & 92.66±0.07 & 34.48±0.02 & 94.18±0.06\\
     & Uniform & 32.46±0.28 & 90.67±0.26 & 34.03±0.11 & 92.40±0.21 & 34.48±0.01 & 94.22±0.04\\
     & Herding & 32.86±0.26 & 91.18±0.48 & 33.99±0.20 & 92.52±0.37 & 34.48±0.02 & 94.20±0.12\\
     & K-Center & 32.92±0.23 & 91.34±0.28 & 34.13±0.17 & 92.68±0.23 & 34.47±0.01 & 94.20±0.07\\
     \midrule
     \rowcolor{gray!12} & IPP=1 & 33.24±0.29 & 91.74±0.32 & 34.45±0.14 & 92.52±0.23 & 34.50±0.01 & 94.20±0.05\\
      \rowcolor{gray!12}\multirow{-2}{*}{Ours} & IPP=5 & \textbf{34.79±0.04} & \textbf{93.16±0.19} & \textbf{35.48±0.05} & \textbf{94.57±0.03} & \textbf{34.52±0.01} & \textbf{94.30±0.01}\\
    \bottomrule
  \end{tabular}
  }
  \label{tab:net_gen}
\end{table}

\noindent \textbf{Cross-Architecture Experiment.} To verify that our distilled dataset can be effectively utilized across different downstream models, we design cross-architecture experiments. Specifically, the dataset is distilled using SRCNN~\cite{dong2015image} and then used to train and test on different downstream architectures, including SRCNN, EDSR~\cite{lim2017enhanced}, and SCTSRN~\cite{yu2017computed}, under the NRI=10 setting. Quantitative results are shown in Tab.~\ref{tab:net_gen}. Our method achieves the best performance when using EDSR, reaching a PSNR of 35.48dB and an SSIM of 94.57, even outperforming the results obtained with SRCNN. This indicates that the distilled dataset effectively captures information from the raw dataset, without requiring the distillation and downstream networks to share the same architecture. These experiments demonstrate that our method has strong generalization capability and our distilled dataset can be effectively applied across different architectures.

\begin{figure}[!t]
  \centering
  \includegraphics[width=\linewidth]{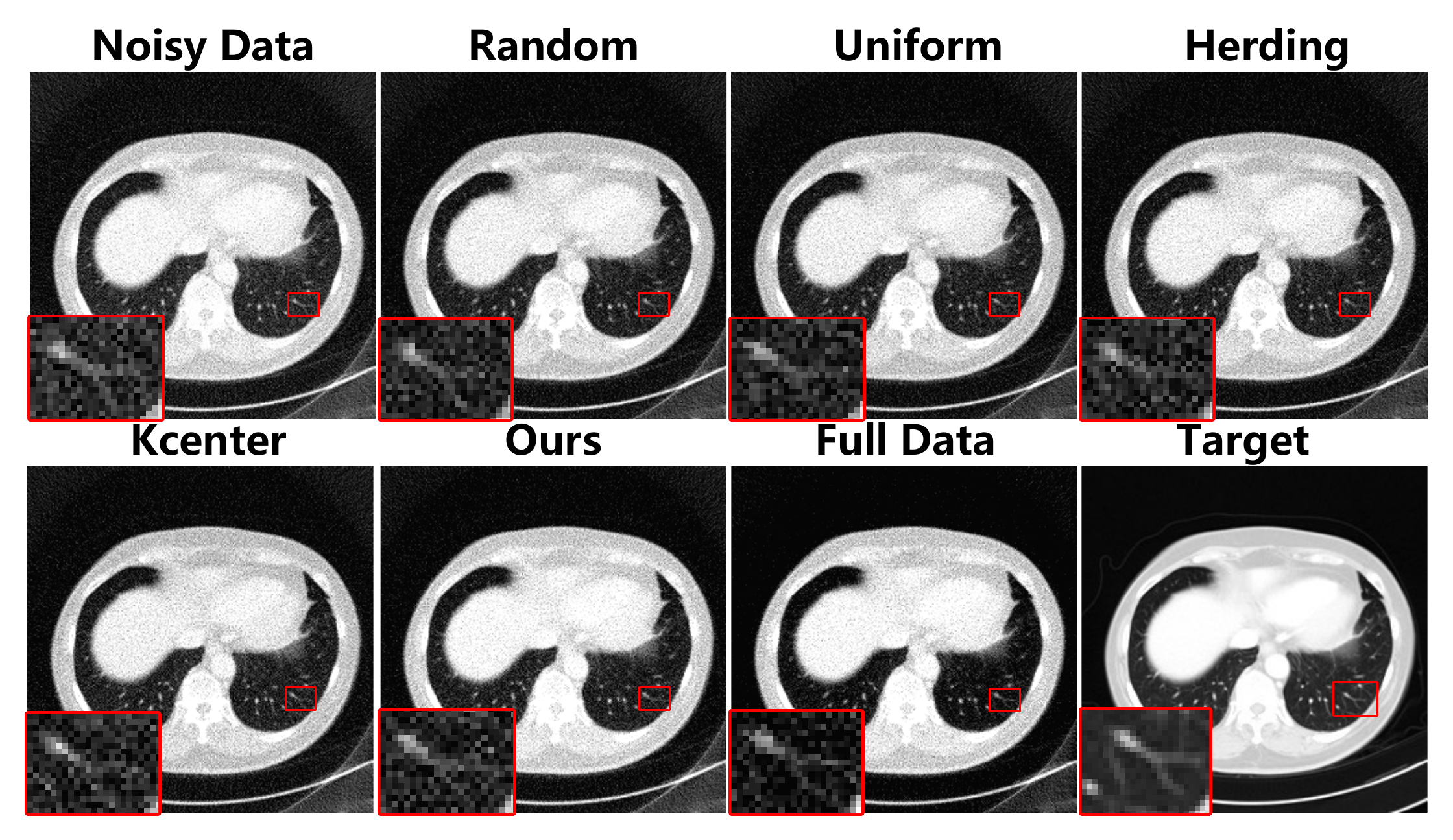}
  \caption{Qualitative results of different algorithms.}
  \label{fig:results_ldct}
\end{figure}

\noindent \textbf{Image Restoration Tasks.} To verify that our method is not only applicable to the super-resolution task, we test it on the CT restoration task. Specifically, we employ the classical REDCNN~\cite{chen2017low} for LDCT restoration. As shown in Tab.~\ref{tab:restoration_result}, our method achieves the best restoration performance, comparable to training with the full dataset. Additionally, we present our quantitative results in Fig.~\ref{fig:results_ldct}. It can be observed that the denoised results obtained using our distilled dataset exhibit higher image quality and better preservation of fine details.

\begin{table}[!t]
  \centering
  \small
  \caption{Quantitative restoration results of different methods trained on REDCNN for the CT modality.}
  \resizebox{\columnwidth}{!}{
  \begin{tabular}{llllll}
    \toprule
     & & \multicolumn{2}{c}{NRI=5} & \multicolumn{2}{c}{NRI=10}\\
    \cmidrule(r){3-6}
    \multicolumn{2}{c}{Methods} & PSNR~(dB)$\uparrow$ & SSIM$\uparrow$ & PSNR~(dB)$\uparrow$ & SSIM$\uparrow$\\
    \midrule
    \multicolumn{2}{c}{Full data}& \multicolumn{2}{c}{PSNR=28.90±0.01} & \multicolumn{2}{c}{SSIM=58.08±0.05}\\
    \midrule
    \multirow{5}{*}{Base} & Random & 28.06±0.01 & 56.54±0.06 & 28.10±0.02 &56.75±0.12\\
     & Uniform & 28.04±0.01 & 57.06±0.06 & 28.07±0.00 & 57.06±0.06\\
     & Herding & 28.05±0.01 & 56.74±0.07 & 28.10±0.02 & 56.75±0.17\\
     & K-Center & 28.05±0.00 & 56.56±0.04 & 28.11±0.05 & 56.78±0.32\\
    \midrule
    \rowcolor{gray!12} & IPP=1  & 28.00±0.07 & 57.20±0.62 & 28.09±0.03 & 56.36±0.42\\
    \rowcolor{gray!12}\multirow{-2}{*}{Ours} & IPP=5  & \textbf{28.38±0.08} & \textbf{57.45±0.63} & \textbf{28.38±0.14} & \textbf{57.43±0.94}\\
    \bottomrule
  \end{tabular}
  }
  \label{tab:restoration_result}
\end{table}

\noindent \textbf{Large-Scale Dataset Experiment.} To further evaluate the effectiveness of the proposed method on a large-scale dataset, we conduct experiments using data from 48 patients in the public CT dataset~\cite{mccollough2016tu}, while reserving data from two additional patients for testing. Since computing the loss over all patients before backpropagation is computationally expensive, we accumulate gradients over every 5 patients and then perform a backpropagation step, iteratively updating the synthetic samples until all patients are processed. As shown in Tab.~\ref{tab:multi-patient}, under the same NRI setting, our method achieves a PSNR of 36.08 dB and an SSIM of 94.94, which is closest to the performance obtained with full-dataset training. This experiment demonstrates that our method has the potential to be applied to larger-scale datasets.

\begin{figure}[!t]
  \centering
  \includegraphics[width=\linewidth]{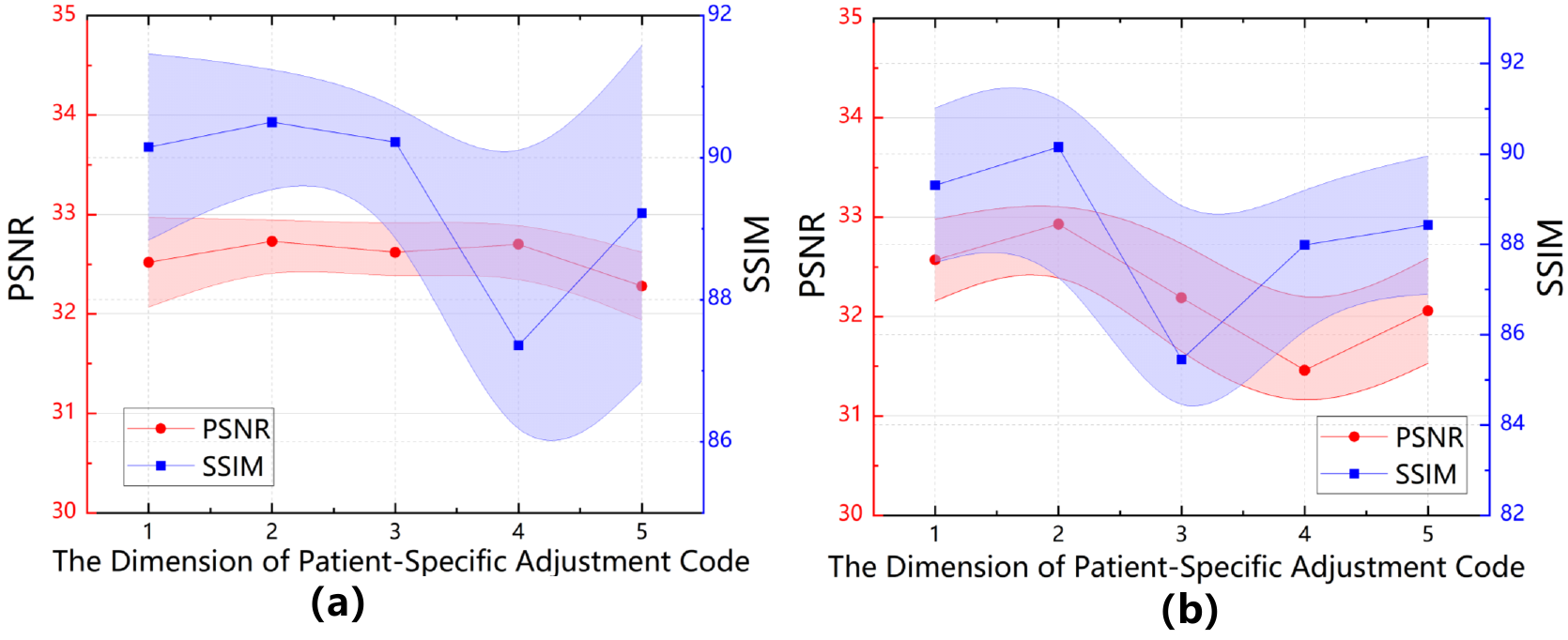}
   \caption{Parameter validation experiments for the dimension of the patient-specific adjustment code. (a) and (b) denote the results under NRI=5 and NRI=10, respectively.}
   \label{fig:dynamic}
\end{figure}

\begin{table}[!t]
  \centering
  \small
  \caption{Quantitative super-resolution results of different methods based on the large-scale dataset.}
  \resizebox{\columnwidth}{!}{
  \begin{tabular}{llllll}
    \toprule
     & & \multicolumn{2}{c}{NRI=5} & \multicolumn{2}{c}{NRI=10} \\
    \cmidrule(r){3-6}
    \multicolumn{2}{c}{Methods} & PSNR~(dB)$\uparrow$ & SSIM$\uparrow$ & PSNR~(dB)$\uparrow$ & SSIM$\uparrow$\\
    \midrule
    \multicolumn{2}{c}{Full Data} & \multicolumn{2}{c}{PSNR=36.95±0.02} & \multicolumn{2}{c}{SSIM=95.58±0.02}\\
    \midrule
    \multirow{5}{*}{Base} & Random & 31.27±0.23 & 87.79±0.57 & 32.65±0.34 & 91.64±0.29\\
     & Random* & 31.81±0.18 & 89.26±0.67 & 31.47±0.12 & 88.45±0.74\\
     & Uniform & 31.21±0.15 & 87.22±0.98 & 32.95±0.12 & 91.54±0.75\\
     & Herding & 31.55±0.20 & 88.03±1.04 & 32.74±0.22 & 91.66±0.29\\
     & K-Center & 31.75±0.45 & 89.13±0.84 & 32.76±0.28 & 91.74±0.36\\
    \midrule
     \rowcolor{gray!12} & IPP=1 & 34.80±0.39 & 92.37±2.17 & 34.91±0.18 & 92.91±1.54\\
     \rowcolor{gray!12}\multirow{-2}{*}{Ours} & IPP=5 & \textbf{36.08±0.06} & \textbf{94.94±0.05} & \textbf{36.07±0.07} & \textbf{95.00±0.08}\\
    \bottomrule
  \end{tabular}
  }
  \label{tab:multi-patient}
\end{table}

\subsection{Ablation Study and Analysis}
\label{exp:ablation}
\noindent \textbf{Hyper-parameter Verification.} To evaluate the impact of the patient-specific adjustment code dimensions, we conduct hyperparameter validation experiments. As shown in Fig \ref{fig:dynamic}, the left figure corresponds to NRI = 5, and the right figure corresponds to NRI = 10. For each plot, the horizontal axis denotes the dimension of the patient-specific adjustment code, while the vertical axis shows the PSNR and SSIM values. From the results, we observe that setting the patient-specific adjustment code to be 2-dimensional yields the best and most stable performance. Therefore, we empirically recommend a dimensionality of 2.

\noindent \textbf{Ablation Experiment.} 
Our proposed SPG consists of personalization and pixel-level fidelity preservation steps, and we conduct experiments to verify their effectiveness. Specifically, ``Ours$\dagger$" refers to our method initialized with random noise, while ``Ours$\ddagger$" denotes our method without the pixel-level fidelity preservation step.
As shown in Tab.~\ref{tab:abl}, we observe that incorporating SPG not only reduces storage requirements but also improves performance. Remarkably, our method can compress the dataset by nearly 99\% while maintaining performance comparable to that of the full dataset. Additional ablation studies are provided in the \textbf{\textit{Appendix}}.

\begin{table}[!t]
  \centering
  \small
  \caption{Ablation study of our method. Below the storage size, we report the reduction rate, which is calculated by dividing the compressed storage size by the original storage size.}
  \resizebox{\columnwidth}{!}{
  \begin{tabular}{lllllll}
    \toprule
    \multirow{2}{*}{Methods} & \multicolumn{3}{c}{IPP=1} & \multicolumn{3}{c}{IPP=5}\\
    \cmidrule(r){2-4} \cmidrule(r){5-7}
     & PSNR~(dB)$\uparrow$ & SSIM$\uparrow$ & Storage$\downarrow$ & PSNR~(dB)$\uparrow$ & SSIM$\uparrow$ & Storage$\downarrow$\\
    \midrule
    Full Data & \multicolumn{2}{c}{PSNR=36.10±0.02} & \multicolumn{2}{c}{SSIM=94.86±0.03} & \multicolumn{2}{c}{Storage=39.9MB} \\
    \midrule
    Ours$\dagger$ & 11.54 & 45.79 & 817KB & 11.30 & 45.55 & 4085KB\\
    & \footnotesize ±0.28 & \footnotesize ±0.36 & \footnotesize $-98.00\%$ & \footnotesize ±0.19 & \footnotesize ±0.23 & \footnotesize $-90.00\%$\\
    Ours$\ddagger$ & 16.95 & 51.57 & 410KB & 30.55 & 88.11 & 412KB \\
    & \footnotesize ±1.40 & \footnotesize ±1.83 & \footnotesize $-98.99\%$ & \footnotesize ±0.06 & \footnotesize ±0.16 & \footnotesize $-98.99\%$ \\
    Ours NRI=5 & 32.73 & \textbf{90.50} & 410KB & 34.45 & 92.07 & 412KB \\
     & \footnotesize ±0.22 & \footnotesize ±0.78 & \footnotesize $-98.99\%$ & \footnotesize ±0.20 & \footnotesize ±0.93 & \footnotesize $-98.99\%$ \\
    Ours NRI=10 & \textbf{32.93} & 90.15 & 819KB & \textbf{34.57} & \textbf{92.83} & 822KB \\
    & \footnotesize ±0.28 & \footnotesize ±1.90 & \footnotesize $-97.99\%$ & \footnotesize ±0.14 & \footnotesize ±0.28 & \footnotesize $-97.98\%$\\
    \bottomrule
  \end{tabular}
  }
  \label{tab:abl}
\end{table}

\begin{figure}[!t]
    \centering
    \includegraphics[width=0.9\linewidth]{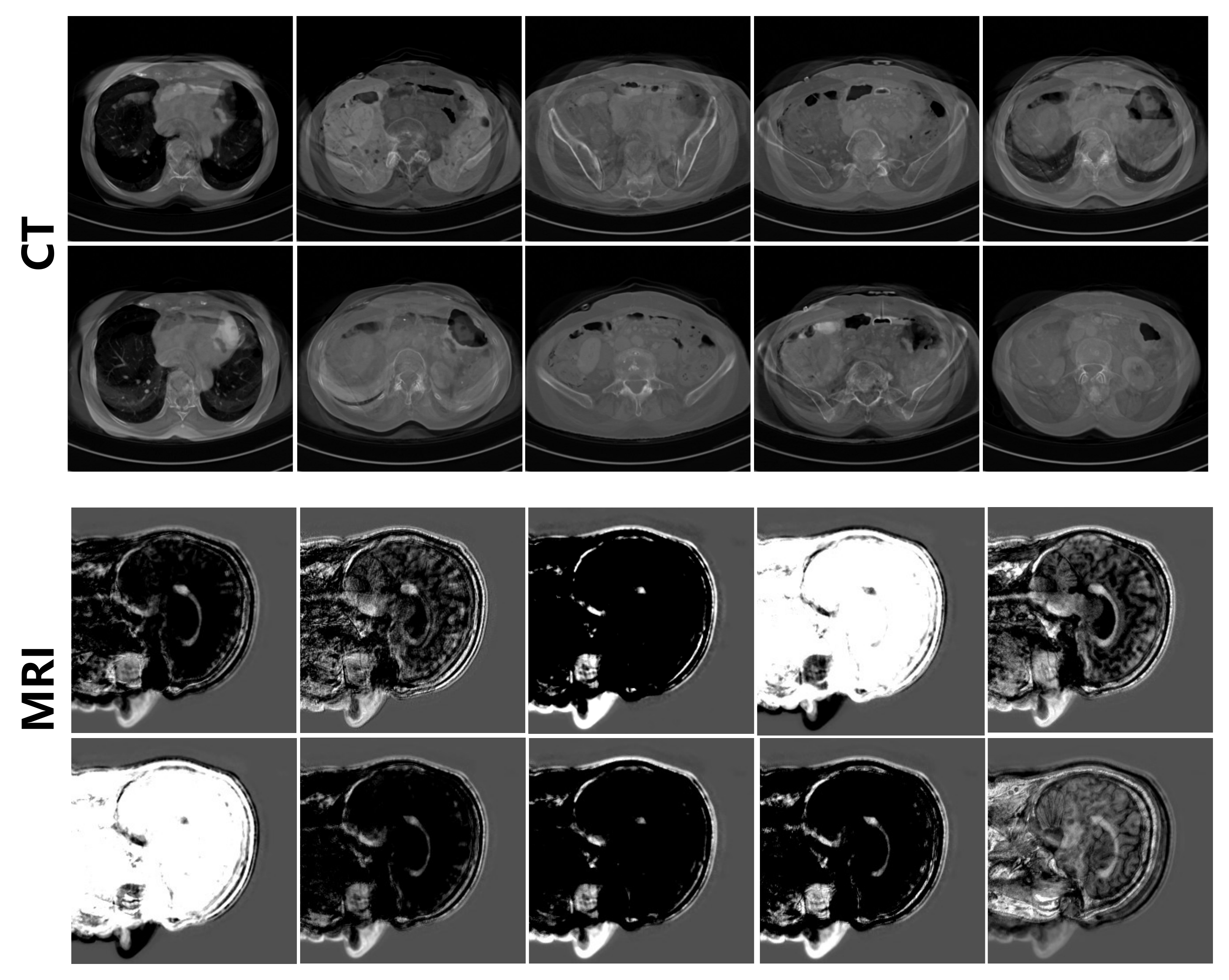}
    \caption{Distillation results of our methods. Each image represents the synthetic data of a patient.}
    \label{fig:syn_data}
\end{figure}

\noindent \textbf{Visualization of the Distilled Data.} We present the synthetic data distilled by our method. When NRI = 10 and IPP = 1, the synthesized samples for both CT and MRI modalities are shown in Fig.~\ref{fig:syn_data}. It can be observed that the synthetic data integrate anatomical characteristics from multiple images, while still preserving variations across different patients, thereby aiding in the preservation of patient privacy. 

\begin{figure}[!t]
  \centering
  \includegraphics[width=\linewidth]{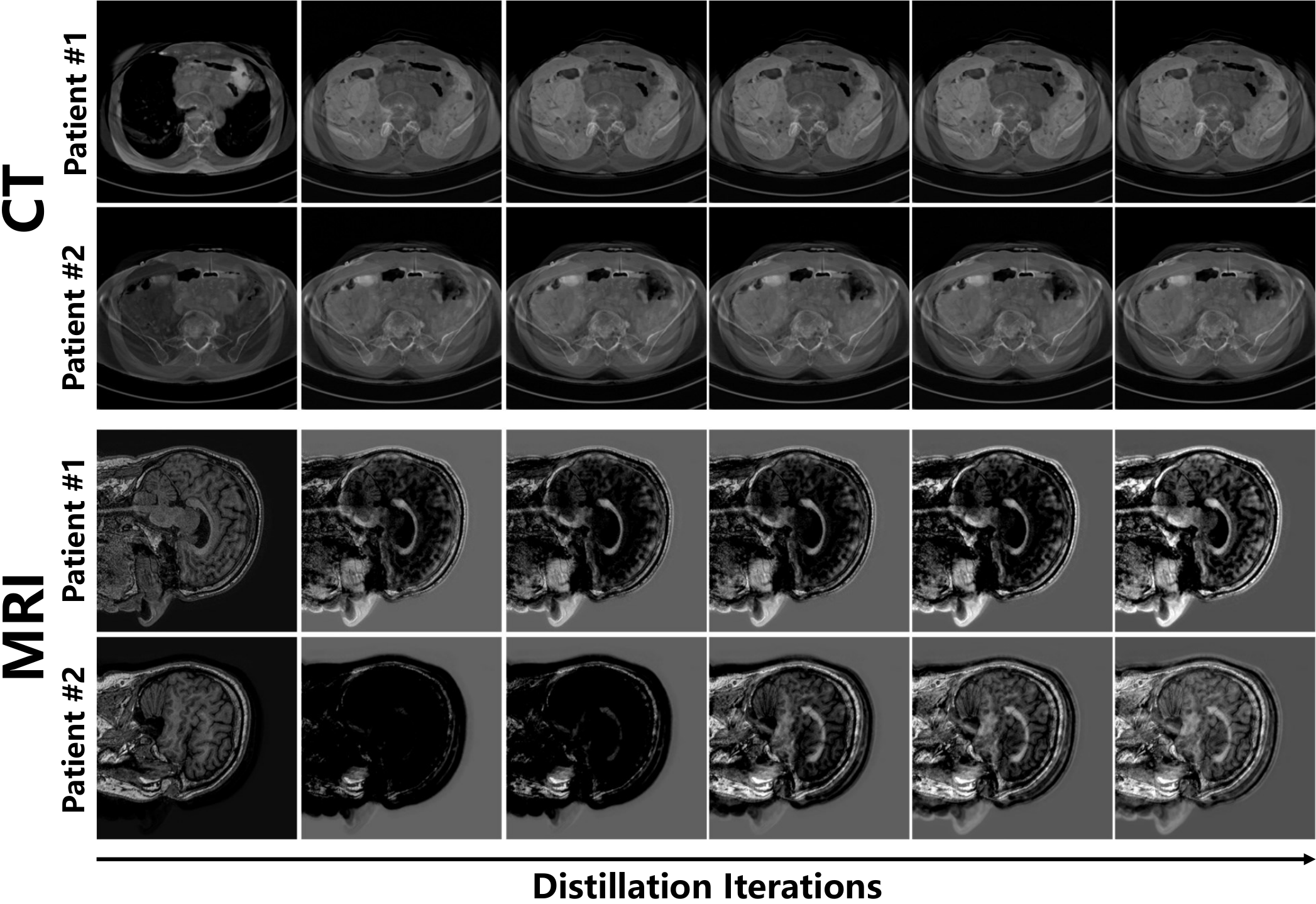}
  \caption{Visualization of the distillation process.}
  \label{fig:distillate_process}
\end{figure}

\noindent \textbf{Visualization of the Distillation Process.} We visualize the synthetic images across different training iterations to illustrate their evolution. Under the setting of NRI = 10 and IPP = 1, the distilled results at different iterations are shown in Fig.~\ref{fig:distillate_process}. As observed, the synthetic images evolve from their initial state and converge, demonstrating that our method effectively injects patient-specific information into the distilled data. Notably, throughout the entire process, no patient data is explicitly transmitted to the user, thereby preserving privacy. Additional privacy-related experiments can be found in our \textit{\textbf{Appendix}}.

\section{Conclusion}
\label{sec:conclusion}
We propose the first low-level dataset distillation framework for medical image enhancement. Our approach leverages anatomical similarity to construct a shared prior, which is then personalized through a learnable modulation module to produce patient-specific distilled data. Task-specific high–low quality pairs and patient-wise gradient alignment ensure that both task-relevant information and patient-wise anatomical information are encoded without directly sharing raw data. Experiments across multiple modalities, tasks, and network architectures show that our method achieves strong enhancement performance with a compact synthetic dataset, offering an efficient and privacy-preserving solution for medical image enhancement.
For future work, we plan to extend the proposed low-level dataset distillation framework to additional imaging modalities and broader low-level tasks, further validating its applicability in diverse scenarios such as continual learning, neural architecture search, and federated learning.
{
    \small
    \bibliographystyle{unsrt} % plainnat
    \bibliography{main}
}

% \input{sec_new/X_suppl}

\end{document}